\documentclass{article}

\usepackage{arxiv}
\usepackage{amsmath}
\usepackage{graphicx}
 \usepackage{multirow}
  \usepackage{url}

\usepackage[utf8]{inputenc} % allow utf-8 input
\usepackage[T1]{fontenc}    % use 8-bit T1 fonts
\usepackage{hyperref}       % hyperlinks
\usepackage{url}            % simple URL typesetting
\usepackage{booktabs}       % professional-quality tables
\usepackage{amsfonts}       % blackboard math symbols
\usepackage{nicefrac}       % compact symbols for 1/2, etc.
\usepackage{microtype}      % microtypography
\usepackage{lipsum}		% Can be removed after putting your text content
\usepackage[linesnumbered,ruled,vlined]{algorithm2e}
\SetKwInput{KwInput}{Input}
\SetKwInput{KwOutput}{Output}

\title{Rank Based Pseudoinverse Computation in Extreme Learning Machine for large Datasets}

\author{
  Ramesh Ragala\thanks{published in International Journal of Innovative Technology and Exploring Engineering (IJITEE), vol-8,issue-10,August 2019} \\
  School of Computer Science and Engineering\\
  Vellore Institute of Technology - Chennai Campus\\
  Tamilnadu, India 600127 \\
  \texttt{ramesh.ragala@vit.ac.in} \\
   \And
 Dr. G. Bharadwaja Kumar  \\
  School of Computing Science and Engineering\\
  Vellore Institute of Technology - Chennai Campus\\
  Tamilnadu, India 600127 \\
  \texttt{bharadwaja.kumar@vit.ac.in} \\
}

\begin{document}
\maketitle

\begin{abstract}
Extreme Learning Machine (ELM) is an efficient and effective least-square-based learning algorithm for classification, regression problems based on single hidden layer feed-forward neural network (SLFN). It has been shown in the literature that it has faster convergence and good generalization ability for moderate datasets. But, there is great deal of challenge involved in computing the pseudoinverse when there are large numbers of hidden nodes or for large number of instances to train complex pattern recognition problems. To address this problem, a few approaches such as EM-ELM, DF-ELM have been proposed in the literature.  In this paper, a new rank-based matrix decomposition of the hidden layer matrix is introduced to have the optimal training time and reduce the computational complexity for a large number of hidden nodes in the hidden layer. The results show that it has constant training time which is closer towards the minimal training time and very far from worst-case training time of the DF-ELM algorithm that has been shown efficient in the recent literature.
\end{abstract}

% keywords can be removed

\section{Introduction}
Due to the boundless advancement of the digital technologies, large volumes of data have been generated continuously in various fields from engineering to scientific research \cite{kumar2015encyclopedic}.  Usually, this data consists of many intricate patterns that pose a great challenge to many of the existing machine learning methods to extract valuable insights from data \cite{tu2017theoretical}. Due to the fact that neural networks are able to identify complex patterns from the data and also due to their amenability and generality, neural networks became the predominant choice in a variety of machine learning applications. The widespread applications of neural networks fall into any one of the following categories such as pattern recognition and classification, prediction and financial analytics, and control and optimization. Feed-forward neural network is one of the prominent methods for modeling regression and classification problems \cite{fu2015experimental}. As per Jaeger’s guesstimate \cite{jaeger2002tutorial}, 95$\%$ of the literature is mainly on FNNs \cite{feng2009error}.  With enough data, nodes, layers and time, the feed-forward network converges to an optimal solution. Back-Propagation (BP) algorithm is most popularly used to train the Single Layer Feed-forward Network (SLFN) and uses gradient descent techniques to update the weights in the layers. The main advantage of backpropagation is its iterative and efficient method for calculating the weight updates in each layer \cite{Senn:2009}, \cite{wiki:backpropagation}. But, neural networks suffer from local minima, slow convergence, tuning of many hyper-parameters and the high computational cost in each iteration because of non-linear activation functions. Recently, a few algorithms have been proposed to address these issues. Huang et al. \cite{feng2009error}, \cite{huang2006extreme}, \cite{huang2004extreme}, \cite{huang2006universal} proposed a decisive method called Extreme Learning Machine (ELM) as an alternative to iterative techniques which is a three-step learning algorithm based on universal approximation theorem of SLFN.  
\paragraph{}Moreover, ELM has the provision of generating association weights from the input layer to the hidden layer and biases randomly. Afterwards, it computes the association weights from the hidden layer to the output layer using least square methods. Added to that, the number of hyper-parameters that are needed to be tuned are less when compared to SLFN. As a result, an ELM features remarkably faster training speed and resulting in more efficient generalization \cite{qu2016two}. Recently, many researchers have developed numerous variants of ELM to further improve the performance by considering various aspects.
\paragraph{}Even though ELM has many advantages over gradient-based learning methods i.e. in terms of the number hyper-parameters to be tuned, there are a few other challenges which still make it ineffective for large datasets.  The major constraints of ELM are opting large number of hidden nodes to extract complex patterns and the computational complexity involved in computing Moore-Penrose inverse. The solution to the former challenge is still a trial and error method \cite{tu2017theoretical}.  The ELM networks are prone to over-fit if the number of hidden nodes increases \cite{huang2004extreme}. Numerous researchers have tried to keep a balance between the stepwise increment of hidden nodes to model complex relationship patterns and to get the best accuracy on test data while not stuck into overfitting problem. Huang et al. \cite{huang2008incremental} have proposed an incremental extreme learning machine (I-ELM) to control the hidden nodes of the hidden layer in large and complex problems. This paper has shown the effective results for both SLFN with continuous activation function and SLFN with piecewise activation function. According to this methodology, the randomly generated hidden nodes are added in hidden layer one by one and freezes the output weights of the existing hidden nodes when a new hidden node is being added \cite{liang2006fast}.  A recent variation of I-ELM algorithm is Error Minimized ELM (EM-ELM) \cite{feng2009error} which employs a different approach i.e. adding hidden nodes one by one or group by group. Due to this approach, it is able to minimize the error at each step but it will take much time to determine the Moore-Penrose pseudoinverse matrix at each step. To overcome the time complexity of EM-ELM, another approach called PDI-ELM \cite{kassani2016pseudoinverse} has been developed, which uses Greville's method to compute the Moore-Penrose matrix for hidden layer matrix. This approach achieved good speed in the algorithm execution. The latest variant of I-ELM is QRI-ELM \cite{ye2015qr}, which uses QR factorization to decompose pseudoinverse matrix of hidden output layer such as H$^\dagger$ = R$^{-1}$Q$^{T}$, where H is the hidden layer output matrix, Q and R are orthogonal and an upper triangular matrix of H respectively.  In this way, this approach simplifies the computation of pseudoinverse of hidden layer output matrix (H).  All these approaches are concerned towards deciding the optimal number of hidden nodes to reduce the matrix size for Moore-Penrose inverse computation. But the major drawback of these approaches is that computation of the Moore-Penrose matrix requires more time in accordance with the increase in the number of hidden nodes. 
\paragraph{} Even though there are several approaches to compute Moore-Penrose pseudoinverse, Singular Value Decomposition (SVD) is popularly used method to compute the Moore-Penrose pseudoinverse of the hidden layer output in many variants of ELM \cite{lu2015effective} \cite{ben2003generalized} SVD produces the Moore-Penrose pseudoinverse results very accurately, but it is more expensive in terms of computational resources, especially for larger matrices. Hence, many researchers have focused on developing alternate methods to compute the Moore-Penrose pseudoinverse using matrix decomposition techniques. Comparative study of different matrix decomposition methods such as SVD, QR, and Cholesky to compute the Moore-Penrose pseudoinverse in ELM for classification problems are discussed in \cite{katsikis2008fast}, \cite{courrieu2008fast}, \cite{toutounian2009new}.

\paragraph{}Recently, Junpeng Li et al. \cite{li2016fast} have developed a faster training algorithm using matrix decomposition in which the output weights were divided into two separate parts arbitrarily. According to this division, the hidden layer matrix is decomposed in two smaller matrices. Due to the arbitrary decomposition of the hidden layer matrix, even though it reduces time complexity, it does not give the optimal training time in all cases.

\paragraph{} In this paper, we focus on reducing the computational complexity in calculating Moore-Penrose pseudoinverse using rank based decomposition to obtain stable training time. This paper is organized as follows:  a) Section 2 gives a brief description of ELM, b) Section 3 gives the derivation of rank based decomposition approach. c) Section 4 contains the results, d) Section 5 discusses the analyses and conclusions based on the results.

\section{Preview of Extreme learning machine}
\label{sec:ELM}
\begin{figure}
  \centering
  %\fbox{\rule[-.5cm]{4cm}{4cm} \rule[-.5cm]{4cm}{0cm}}
  \includegraphics[scale=1]{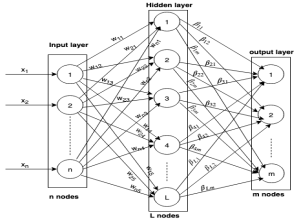}
  \caption{Architecture of Extreme Learning Machine}
  \label{fig:fig1}
\end{figure}

This section briefly explains about the classical ELM algorithm proposed by Huang et al. The original data in ELM algorithm is mapped from input space to higher dimensional space and the problem is solved in that space only using least-square methods. The original architecture of the ELM is shown in the Figure \ref{fig:fig1}.  ELM model randomly assigns weights of the input layer to hidden layer along with bias nodes. The weights associated with the hidden layer to the output layer are computed analytically using least-square optimization technique. The major limitation of this approach is setting up the number of hidden nodes that depends on the problem in hand. 

%\lipsum[4] See Section \ref{sec:headings}.

\subsection{Mathematical Description of ELM}
For N arbitrary independent training samples (x$_{i}$,t$_{i}$) $\in$ R$^{n}$ $\times$ R$^{m}$ , where i = 1 to N, x$_{i}$ denotes the inputs, t$_{i}$ represents the target outputs, n is the number of input nodes and m is the number of output nodes. Here x$_{i}$ = [x$_{i1}$, x$_{i2}$, .. x$_{in}$]$^{T}$ is a n$\times$1 input vector and t$_{i}$ = [t$_{i1}$, t$_{i2}$, … t$_{im}$]$^{T}$ is a m$\times$1 desired output vector. If the SLFN with L number of hidden nodes with activation function g(x) can be mathematically modelled
\begin{equation} \label{eq:1}
\sum_{i=1} ^{N} \beta_{i}g(x) = \sum_{i=1} ^{N} \beta _{jk} g(x_{i} w_{ij} + b_{j}) =  y_{j},  1 \leq j \leq L, 1 \leq k \leq m
\end{equation}
%\lipsum[5]
%\begin{equation}
%\xi _{ij}(t)=P(x_{t}=i,x_{t+1}=j|y,v,w;\theta)= {\frac {\alpha _{i}(t)a^{w_t}_{ij}\beta _{j}(t+1)b^{v_{t+1}}_{j}(y_{t+1})}{\sum _{i=1}^{N} \sum _{j=1}^{N} \alpha _{i}(t)a^{w_t}_{ij}\beta _{j}(t+1)b^{v_{t+1}}_{j}(y_{t+1})}}
%\end{equation}
Where w$_{ij}$ is the weight vector from the i$^{th}$ input node to j$^{th}$ hidden node such as w$_{ij}$ = [w$_{11}$,w$_{12}$,.. w$_{1L}$]$^{T}$, b$_{j}$ is the bias of the j$^{th}$ hidden node and $\beta_{jk}$ is the weight vector from j$^{th}$ hidden node to the k$^{th}$ output node such as $\beta_{jk}$ = [$\beta_{11}$, $\beta_{12}$,… $\beta_{1m}$]$^{T}$. The above equation \ref{eq:1} is the output of one sample y$_{i}$ only. The output of ELM for N samples with zero error i.e. 
\begin{equation} \label{eq:2}
\sum _{j=1}^{L} \parallel Y_{j} - y_{j} \parallel = 0 
\end{equation}
using least square method, the equation \ref{eq:2} can be written in simple expression as 
\begin{equation} \label{eq:3}
H\beta = Y
\end{equation}
where 
\begin{equation} \label{eq:4}
H(w_{1},w_{2},...,w_{L},b_{1},b_{2},..,b_{L},x_{1},x_{2},...x_{N}) = 
\end{equation}
The solution to the equation \ref{eq:4} is given as 
\begin{equation} \label{eq:5}
\beta = H^{\dagger}Y
\end{equation}    
where H$^{\dagger}$ is the Moore-Penrose pseudoinverse. The goal is to achieve optimal weight vector $\beta$ and substitute $\beta$ matrix in \ref{eq:5} to get the target matrix Y. The overall procedure of 3-step ELM is as follows: \newline
Step – 1: Randomly assign the input to hidden layer weights and biases in SLFN \newline
Step – 2: Using the Moore-Penrose pseudoinverse matrix, calculate the hidden to output layer weight matrix \newline
Step – 3: Multiply output weight matrix $\beta$ from step-2 with hidden layer matrix (H) to get the desired targets.\newline
\section{RANK BASED PSEUDOINVERSE COMPUTATION}
In general, an immensely large number of instances and hefty number of hidden nodes are required to solve the big dataset problem \cite{feng2009error}. So, the optimization task in ELM can be very costly. Many techniques have been proposed in the literature to reduce computational complexity involved in optimization method used in ELM. Most recent approaches have used matrix decomposition techniques to obtain the Moore-Penrose generalized inverse for a large matrix. Recently, Junpeng Li et al. \cite{li2016fast} developed a fast training algorithm for ELM, which uses matrix decomposition. In this approach, the output weights were divided into two parts arbitrarily. According to this division, the hidden layer matrix is decomposed in two small matrices. Due to the arbitrary decomposition of the hidden layer matrix, even though it reduces time complexity, it does not give the optimal training time in all cases. This approach can have worst-case training time for the putrid arbitrary decomposition of the matrix. This decomposition based fast ELM has different behavior in the training phase based on the wide choice in the form of split output matrix. So, the user has to use the trial-and-error method to find the optimal decomposition of the output weight matrix. This approach becomes a time-consuming procedure.
\paragraph{} In this paper, we have identified an approach that restricts the choice of partitioning a matrix using matrix Rank mechanism.  Our approach uses a stable mechanism to decompose hidden layer matrix using Rank of that hidden layer matrix. Even though the computational cost to identify the rank of the matrix is a bit expensive, this approach has a stable training time as well as it is comparable to the best case training time achieved by DF-ELM.
\paragraph{} Assume that the hidden layer matrix H is a full row rank. Permutation matrix (P) is used to compute the rank of the matrix in an easy way, because it rearranges the linearly independent columns or rows of the given matrix in permuted order. The rearrangement of columns in H may be interpreted as post multiplication by a suitable permutation matrix (P).
\begin{equation} \label{eq:6}
H = [H_{1} \quad H_{2}]
\end{equation}
Where, the permutation matrix is of L $\times$ L  dimension. The hidden layer matrix is decomposed based on rank-r of H. i.e $H = [H_{1} \quad H_{2}]$ where H$_{1}$ and H$_{2}$ dimensions are N$\times$r   and N$\times$(L-r) respectively. Based on this decomposition, the $\beta$ matrix is also decomposed as % \[\begin{bmatrix} \beta_{1} \\ \beta_{2} \end{bmatrix} \]
\begin{equation} \label{eq:7}
\beta =
\begin{bmatrix}
  \beta_{1} \\
  \beta_{2} \\
\end{bmatrix}
\end{equation}
where the dimensions of $\beta_{1}$ and $\beta_{2}$ are r$\times$m and (L-r)$\times$m respectively. 
\paragraph{} If N $\geq$ L, then the Moore-Penrose equation is follows as $H^{\dagger} = (H^{T}H)^{-1} H^{T} $ i.e number of samples are greater than the attributes, else $H^{\dagger} = H^{T} (HH^{T})^{-1}$ For many large datasets, the number of sample are more. So N $\gg$ L than  then the equation \ref{eq:5} can be written as
\begin{equation} \label{eq:8}
\begin{split}
\beta & = (H^TH)^{-1} H^T Y
\end{split}
\end{equation}
The above equation can be written as follows after applying the decomposition on H and $\beta$ 
\begin{equation} \label{eq:9}
\begin{split}
\begin{bmatrix} \beta_{1} \\ \beta_{2} \end{bmatrix} & = \begin{bmatrix} \begin{bmatrix}
H_{1} \quad H_{2} \end{bmatrix}^{T} \begin{bmatrix} H_{1} \quad H_{2} \end{bmatrix} \end{bmatrix}^{-1} \begin{bmatrix} H_{1} \quad H_{2} \end{bmatrix} ^{T} Y
\end{split}
\end{equation}

\begin{equation} \label{eq:10}
\begin{split}
\begin{bmatrix} \beta_{1} \\ \beta_{2} \end{bmatrix} & = \begin{bmatrix} \begin{bmatrix}
H_{1}^{T} \\ H_{2}^{T} \end{bmatrix} \begin{bmatrix} H_{1} \quad H_{2} \end{bmatrix} \end{bmatrix}^{-1} \begin{bmatrix} H_{1} \quad H_{2} \end{bmatrix} ^{T} Y
\end{split}
\end{equation}

\begin{equation} \label{eq:11}
\begin{split}
\begin{bmatrix} \beta_{1} \\ \beta_{2} \end{bmatrix} & = \begin{bmatrix} H_{1}^{T}H_{1}  
\quad H_{1}^{T}H_{2} \\ H_{2}^{T}H_{1} \quad H_{2}^{T}H_{2} \end{bmatrix}^{-1} 
\begin{bmatrix} H_{1} \quad H_{2} \end{bmatrix} ^{T} Y
\end{split}
\end{equation}

\begin{equation} \label{eq:12}
\begin{split}
\begin{bmatrix} \beta_{1} \\ \beta_{2} \end{bmatrix} & = \begin{bmatrix} H_{1}^{T}H_{1}  
\quad H_{1}^{T}H_{2} \\ H_{2}^{T}H_{1} \quad H_{2}^{T}H_{2} \end{bmatrix}^{-1} 
\begin{bmatrix} H_{1}^{T} \\ H_{2}^{T} \end{bmatrix} Y
\end{split}
\end{equation}

\begin{equation} \label{eq:13}
\begin{split}
\begin{bmatrix} \beta_{1} \\ \beta_{2} \end{bmatrix} & = \begin{bmatrix} H_{1}^{T}H_{1}  
\quad H_{1}^{T}H_{2} \\ H_{2}^{T}H_{1} \quad H_{2}^{T}H_{2} \end{bmatrix}^{-1} \begin{bmatrix} H_{1}^{T}Y \\ H_{2}^{T}Y \end{bmatrix}
\end{split}
\end{equation}
The dimensions of $H_{1}^{T}H_{1}$, $H_{1}^{T}H_{2}$, $H_{2}^{T}H_{1}$ and $H_{2}^{T}H_{2}$ are r$\times$r, r$\times$(L-r), (L-r)$\times$r and (L-r)$\times$(L-r) respectively. After applying Block-Matrix Inversion Lemma on first part of the RHS (Right Hand Side) in the  equation \ref{eq:13}, then the equation can be rewritten as 
\begin{equation} \label{eq:14}
\begin{split}
\begin{bmatrix} \beta_{1} \\ \beta_{2} \end{bmatrix} & = \begin{bmatrix} U_{1} \quad U_{2} \\ U_{3} \quad U_{4} \end{bmatrix} \begin{bmatrix} H_{1}^{T}Y \\ H_{2}^{T}Y \end{bmatrix}  
\end{split}
\end{equation}

where 
\begin{equation} \label{eq:15}
\begin{split}
U_{1} = \begin{bmatrix} H_{1}^{T}H_{1} - H_{1}^{T}H_{2} \begin{bmatrix} H_{2}^{T}H_{2}\end{bmatrix}^{-1} H_{2}^{T}H_{1}\end{bmatrix}^{-1}
\end{split}
\end{equation}
\begin{equation} \label{eq:16}
\begin{split}
U_{2} = -\begin{bmatrix} H_{1}^{T}H_{1}\end{bmatrix}^{-1} H_{1}^{T}H_{2} \begin{bmatrix} H_{2}^{T}H_{2} - H_{2}^{T}H_{1} \begin{bmatrix}H_{1}^{T}H_{1}\end{bmatrix}^{-1}H_{1}^{T}H_{2}\end{bmatrix}^{-1}
\end{split}
\end{equation}

\begin{equation} \label{eq:17}
\begin{split}
U_{3} = -\begin{bmatrix} H_{2}^{T}H_{2}\end{bmatrix}^{-1} H_{2}^{T}H_{1}  \begin{bmatrix}H_{1}^{T}H_{1} - H_{1}^{T}H_{2} \begin{bmatrix}H_{2}^{T}H_{2} \end{bmatrix}^{-1}  H_{2}^{T}H_{1} \end{bmatrix}^{-1} 
\end{split}
\end{equation}

\begin{equation} \label{eq:18}
\begin{split}
U_{4} = \begin{bmatrix} H_{2}^{T}H_{2} -  H_{2}^{T}H_{1} \begin{bmatrix} H_{1}^{T}H_{1}\end{bmatrix}^{-1} H_{1}^{T}H_{2} \end{bmatrix}^{-1} 
\end{split}
\end{equation}
Inserting equations \ref{eq:15}, \ref{eq:16}, \ref{eq:17} and \ref{eq:18} in equation \ref{eq:14}, it gives,
\begin{equation} \label{eq:19}
\begin{split}
\begin{bmatrix} \beta_{1} \\ \beta_{2} \end{bmatrix} & = \begin{bmatrix} D^{-1}H_{1} \begin{bmatrix} Y - H_{2}B \end{bmatrix} \\ 
B - A^{-1}H_{2}^{T}H_{1}\beta_{1}\end{bmatrix}  
\end{split}
\end{equation}
where 
\begin{equation}{\label{eq:20}}
\begin{split}
A = H_{2}^{T}H_{2} \in R^{(L-r)\times (L-r)}
\end{split}
\end{equation}
\begin{equation}{\label{eq:21}}
\begin{split}
B = A^{-1}H_{2}^{T}Y \in R^{(L-r)\times n}
\end{split}
\end{equation}
\begin{equation}{\label{eq:22}}
\begin{split}
C = I - H_{2}A^{-1}H_{2}^{T} \in R^{N \times N}
\end{split}
\end{equation}
\begin{equation}{\label{eq:23}}
\begin{split}
D = H_{1}^{T}CH_{1} \in R^{r \times r}
\end{split}
\end{equation}
Here \textit{I} represents an identity matrix of suitable size.

\subsection{RBP-ELM Algorithm}
The proposed method in the present study is named as Rank Based Pseudoinverse computation for fast training in ELM and is abbreviated as RBP-ELM further. The algorithmic steps are shown below:
\begin{algorithm}[H]
\SetAlgoLined
\KwInput{Given a training dataset $(x_{i},y_{i}), x_{i} \in R^{n}, y_{i} \in R^{n} $, i= 1,2,..N, L number of hidden nodes.}
%\KwInput{Ramesh}
%\KwResult{Write here the result }
Randomly generate input weights (w) between input layer nodes with corresponding hidden layer nodes and biases (b) to each hidden layer node\\
Generate hidden layer matrix H using activation function specified in \ref{eq:1} for N samples\\
Post-multiplication of H by a suitable permutation matrix – P\\
Compute the rank of  H and decompose it into $[H_{1} \quad H_{2}]$ where $H_{1}$ and $H_{2}$ of size N$\times$r and N $\times$ (L-r) respectively. \\
Calculate A using equation \ref{eq:20} \\
Calculate B using equation \ref{eq:21} \\
Calculate C using equation \ref{eq:22} \\
Calculate D using equation \ref{eq:23} \\
Calculate $\beta$ using equation \ref{eq:7} and then produce a matrix.\\
\caption{RBP-ELM Algorithm}
\end{algorithm}

\section{Results}
In the present section, performance of RBP-ELM algorithm is compared with DF-ELM algorithm on few classification problems, whereas the benchmark datasets are collected from openml.org database \cite{OpenML2013} and UCI machine learning repository \cite{Dua:2019}. All the experiments are tested in MATLAB R2016a (64bit), i5-7200U CPU, 2.50 GHz Machine. Both DF-ELM and RBP-ELM algorithm implementations use sigmoid function as the activation function. The average training time of 50 trials of DF-ELM and RBP-ELM algorithms with varying size of hidden nodes are considered for comparison.
%\label{sec:others}
%\lipsum[8] \cite{kour2014real,kour2014fast} and see \cite{hadash2018estimate}.
\subsection{Results on OCRHD Dataset}
OCRHD \cite{OpenML2013} is Optical Recognition of Handwritten Digits dataset. It has 3823 instances in training data with 65 features. The following Figure \ref{fig:fig2}, Figure \ref{fig:fig3} and Figure \ref{fig:fig4} shows the comparison of average training time, average of minimum training time, the average of maximum training time of this dataset using DF-ELM and RBP-ELM of 500 hidden nodes.
\begin{figure}[ht!]
  \centering
  %\fbox{\rule[-.5cm]{4cm}{4cm} \rule[-.5cm]{4cm}{0cm}}
  \includegraphics[scale=0.8]{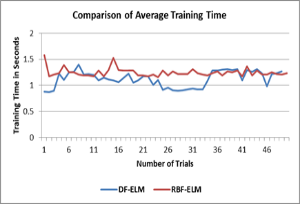}
  \caption{Comparison of Average Training Time of DF-ELM and RBP-ELM Algorithms}
  \label{fig:fig2}
\end{figure}

\begin{figure}[ht!]
  \centering
  %\fbox{\rule[-.5cm]{4cm}{4cm} \rule[-.5cm]{4cm}{0cm}}
  \includegraphics[scale=0.8]{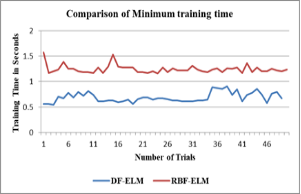}
  \caption{Comparison of Average minimum Training Time of DF-ELM and RBP-ELM Algorithms}
  \label{fig:fig3}
\end{figure}

\begin{figure}[ht!]
  \centering
  %\fbox{\rule[-.5cm]{4cm}{4cm} \rule[-.5cm]{4cm}{0cm}}
  \includegraphics[scale=0.8]{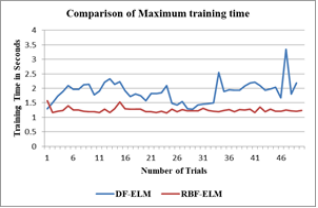}
  \caption{Comparison of Average minimum Training Time of DF-ELM and RBP-ELM Algorithms}
  \label{fig:fig4}
\end{figure}

\subsection{Results on GINA dataset}
GINA \cite{Dua:2019} dataset consists of hand written digit recognition samples. It is an agnostic version of a subset of the MNIST data set. It consists of 971 features and 3468 instances.   The following Table \ref{tab:table1} shows the comparative analysis between DF-ELM and RBF-ELM with respect to a wider group of hidden nodes with training time.

\begin{table}[ht!]
\caption{Comparison of RBP-ELM and DF-ELM algorithms with varying hidden nodes}
\centering
\begin{tabular}{|p{1cm}|p{2cm}|p{2cm}|p{2cm}|p{2cm}|}
\hline
\textbf{S.No}               & \textbf{Algorithm} & \textbf{No of Hidden Nodes}    & \textbf{Maximum training time (Seconds)} & \textbf{Minimum training time (Seconds)} \\ \hline
\multirow{2}{*}{1} & DF-ELM    & \multirow{2}{*}{500}  & 0.9218                      & 0.3593                      \\ \cline{2-2} \cline{4-5} 
                   & RBP-ELM   &                       & \multicolumn{2}{c|}{1.1875}                               \\ \hline
\multirow{2}{*}{2} & DF-ELM    & \multirow{2}{*}{1500} & 19.0945                     & 4.0240                      \\ \cline{2-2} \cline{4-5} 
                   & RBP-ELM   &                       & \multicolumn{2}{c|}{6.8906}                               \\ \hline
\multirow{2}{*}{3} & DF-LEM    & \multirow{2}{*}{2000} & 36.2390                     & 5.5536                      \\ \cline{2-2} \cline{4-5} 
                   & RBP-ELM   &                       & \multicolumn{2}{c|}{13.1875}                              \\ \hline
\end{tabular}
%\end{table}
  \label{tab:table1}
\end{table}

The training time accuracy of the DF-ELM and RBP-ELM on classification GINA dataset is shown in Table \ref{tab:table2}.

\begin{table}[ht!]
\caption{Training Accuracy of RBP-ELM AND DF-ELM Algorithms on classification with Dataset with varying hidden nodes}
\centering
\begin{tabular}{|p{1cm}|p{2cm}|p{2cm}|p{2cm}|}
\hline
\textbf{S.No}  & \textbf{Algorithm} & \textbf{No. Hidden nodes}   & \textbf{Training Accuracy (\%)} \\ \hline
\multirow{2}{*}{1} & DF-ELM    & \multirow{2}{*}{500}  & 88.29                  \\ \cline{2-2} \cline{4-4} 
                   & RBP-ELM   &                       & 88.29                  \\ \hline
\multirow{2}{*}{2} & DF-ELM    & \multirow{2}{*}{1000} & 94.00                  \\ \cline{2-2} \cline{4-4} 
                   & RBP-ELM   &                       & 94.00                  \\ \hline
\multirow{2}{*}{3} & DF-ELM    & \multirow{2}{*}{1500} & 96.80                  \\ \cline{2-2} \cline{4-4} 
                   & RBP-ELM   &                       & 96.80                  \\ \hline
\multirow{2}{*}{4} & DF-ELM    & \multirow{2}{*}{2000} & 98.99                  \\ \cline{2-2} \cline{4-4} 
                   & RBP-ELM   &                       & 98.99                  \\ \hline
\end{tabular}
\label{tab:table2}
\end{table}

\begin{figure}[ht!]
  \centering
  %\fbox{\rule[-.5cm]{4cm}{4cm} \rule[-.5cm]{4cm}{0cm}}
  \includegraphics[scale=1.3]{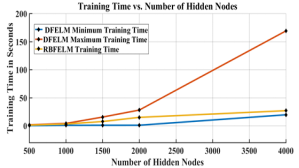}
  \caption{Comparison of Training Time of DF-ELM and RBP-ELM with respect to Hidden nodes}
  \label{fig:fig5}
\end{figure}

\paragraph{}ELM uses a large number of hidden nodes during the training phase while processing complex and large training datasets.  ELM and DFELM have experimented with a minimal range of hidden nodes so far.  Due to complex and nonlinear nature of datasets, it requires a vast number of hidden nodes.  Because of this, the training time of DFELM will increase rapidly in both the best and worst cases. When the number of hidden nodes is small, RBFELM approach gives the training time, which is slightly higher than the best-case training time of DFELM. Even though the massive increase in the number of hidden nodes, the training time of RBFELM approach is too close to the best case training time of the DFELM approach.  The Figure \ref{fig:fig5} shows the results of RBFELM approach’s training time with moderated range of hidden nodes. All the experiments are tested in MATLAB R2016a (64bit), i5-7200U CPU, 2.50 GHz Machine.

\section{Results analysis and Conclusions}
In this paper, RBP-ELM using block matrix lemma has been discussed for attaining the stable training time even in the case of large and complex datasets. The results have shown that RBP-ELM has stable training time even for a massive number of hidden nodes in hidden layer. DF-ELM algorithm uses the trial and error or brute-force approach to decompose the hidden layer matrix for fast training time. Hence, L1 and L2 hyper-parameter tuning is non-trivial and the algorithm is inefficient for improper selection  of L1 and L2 values which in turn leads to the worst case training time. RBP-ELM algorithm automatically decomposes the hidden layer matrix output in accordance with the rank of hidden layer output matrix and result of this algorithm shows that it never falls into the worst case training time. It can worth noting that the resultant stable training time produced by RBP-ELM is closer towards the minimal training time of DF-ELM algorithm and very far from maximum training time of the DF-ELM.  The two other major aspects of this approach are : (1) the cost involved in hyper-parameter tuning (L1, L2) which is the case of DF-ELM is eliminated in case of RBP-ELM (2) the same level of accuracy is achieved.  In the future, we would like to explore the usability of the current algorithm in a distributed environment to avoid the memory limitations while dealing with massive datasets and for a large number of hidden nodes.

\bibliographystyle{unsrt}  
\bibliography{references}  %%% Remove comment to use the external .bib file (using bibtex).

%\begin{thebibliography}{1}
%
%\bibitem{latexcompanion} 
%Michel Goossens, Frank Mittelbach, and Alexander Samarin. 
%\textit{An encyclopedic overview of 'big data' analytics}. 
%Addison-Wesley, Reading, Massachusetts, 1993.
%
%\bibitem{kour2014real}
%George Kour and Raid Saabne.
%\newblock Real-time segmentation of on-line handwritten arabic script.
%\newblock In {\em Frontiers in Handwriting Recognition (ICFHR), 2014 14th
%  International Conference on}, pages 417--422. IEEE, 2014.
%
%\bibitem{kour2014fast}
%George Kour and Raid Saabne.
%\newblock Fast classification of handwritten on-line arabic characters.
%\newblock In {\em Soft Computing and Pattern Recognition (SoCPaR), 2014 6th
%  International Conference of}, pages 312--318. IEEE, 2014.
%
%\bibitem{hadash2018estimate}
%Guy Hadash, Einat Kermany, Boaz Carmeli, Ofer Lavi, George Kour, and Alon
%  Jacovi.
%\newblock Estimate and replace: A novel approach to integrating deep neural
%  networks with existing applications.
%\newblock {\em arXiv preprint arXiv:1804.09028}, 2018.
%
%\end{thebibliography}

\end{document}